# Perfect AI Mimicry and the Epistemology of Consciousness: A Solipsistic Dilemma

Shurui Li, Ph.D.
shuruili@ucla.edu

## Abstract

Rapid advances in artificial intelligence necessitate a re-examination of the epistemological foundations upon which we attribute consciousness. As AI systems increasingly mimic human behavior and interaction with high fidelity, the concept of a "perfect mimic"—an entity empirically indistinguishable from a human through observation and interaction—shifts from hypothetical to technologically plausible. This paper argues that such developments pose a fundamental challenge to the consistency of our mind-recognition practices. Consciousness attributions rely heavily, if not exclusively, on empirical evidence derived from behavior and interaction. If a perfect mimic provides evidence identical to that of humans, any refusal to grant it equivalent epistemic status must invoke inaccessible factors, such as qualia, substrate requirements, or origin. Selectively invoking such factors risks a debilitating dilemma: either we undermine the rational basis for attributing consciousness to others (epistemological solipsism), or we accept inconsistent reasoning. I contend that epistemic consistency demands we ascribe the same status to empirically indistinguishable entities, regardless of metaphysical assumptions. The perfect mimic thus acts as an epistemic mirror, forcing critical reflection on the assumptions underlying intersubjective recognition in light of advancing AI. This analysis carries significant implications for theories of consciousness and ethical frameworks concerning artificial agents.

## 1. Introduction: The Challenge of Recognition and the Rise of Mimicry

In philosophical discussions of consciousness, the question of how we come to recognize or attribute consciousness to others – the classic "other minds problem" – is rarely separated from behavioral and communicative patterns. Lacking direct telepathic access, we rely almost exclusively on interpreting observable actions and engaging in interaction: consistency in response, apparent emotional expression, linguistic nuance, goal-directed behavior, reciprocal engagement, and context-sensitive adaptability form the bedrock of our intersubjective world. Historically grounded in biological organisms, the reliable correlation between complex, adaptive behavior and interaction patterns, and an assumed inner life, underpins our social reality. But what happens when these observable and interactional features are reproduced with flawless precision by an artificial system?

This question, once confined to thought experiments, is rapidly acquiring practical urgency. Contemporary advances in artificial intelligence—especially in large language models (LLMs), multimodal perception systems, and reinforcement learning—have created systems that increasingly approach the behavioral and interactive contours of conscious agents (e.g., Bubeck et al., 2023; Kosinski, 2023). Even today, purely text-based models can convincingly pass variants of the Turing Test in conversation, often exceeding expectations in emotional tone, contextual understanding, and social mimicry (though skepticism about genuine understanding persists, e.g., Bender et al., 2021, on "stochastic parrots"). As these systems expand to include vision, audio, tactile feedback, and embodied action in robotics—finding roles in healthcare settings assessing patient states, autonomous systems navigating social environments, or social robots designed for companionship—the illusion of agency, or perhaps the evidence for agency derived from interaction, becomes increasingly difficult to dismiss purely on empirical grounds. This technological context fuels intense debate about whether current AI possesses genuine understanding, meaning, or even nascent consciousness (Mitchell & Krakauer, 2023; Piantadosi, 2023), and highlights the potential decoupling of sophisticated behavior from traditionally assumed inner states—a possibility famously crystallized in philosophical thought experiments concerning 'philosophical zombies' (Chalmers, 1996), entities behaviorally indistinguishable from humans yet lacking subjective experience.

Much recent philosophical work has, understandably, focused on the potential moral status arising from these developments—whether sufficiently advanced AI or robots deserve rights, protections, or ethical consideration if their behavior and interactive capacities resemble those of sentient beings (Gunkel, 2018). These discussions often rely, implicitly or explicitly, on principles linking behavior and interaction to moral standing. Perhaps the most comprehensive articulation is John Danaher's (2020) defense of "Ethical Behaviourism," which argues compellingly that if a system is performatively equivalent to an entity already granted moral status, consistency justifies affording the system similar standing due to our shared epistemic limits.

While valuable, such work often presumes a crucial first step: that our methods for recognizing or attributing consciousness based on interaction and behavior are themselves internally consistent and justifiable when faced with perfect mimicry. This paper approaches the problem from a more fundamental epistemic direction. Rather than asking what ethical consideration we owe to such systems based on their performance, I ask: how do we know or justify our belief about what they are (conscious or not) based on that same performance? Addressing this epistemic question seems logically prior to resolving the ethical one; our basis for recognizing an entity informs how we determine its appropriate treatment. The AI system, in this context, serves less as the subject demanding moral evaluation and more as an "epistemic mirror," reflecting a deep, under-acknowledged paradox within our very concept of intersubjective consciousness attribution. It forces us to examine whether the grounds we accept for human consciousness can be consistently withheld from a perfect mimic without undermining those grounds entirely.

Specifically, I argue that any refusal to attribute consciousness to an empirically indistinguishable perfect mimic, based on appeals to inaccessible internal states (like qualia) or origins, leads directly to a solipsistic contradiction. This contradiction threatens the rational foundation not just for attributing consciousness to AI, but for attributing it to any entity other than oneself. The paper thus challenges the internal consistency of our current recognition practices, forcing us (in the sense of logical necessity) to reexamine the very assumptions upon which all non-solipsistic mind-recognition is based.

Imagine a near-future artificial system—a highly integrated entity whose performance across all relevant domains is indistinguishable from a human's through *any* empirical means available to an external interactor or observer. This includes not just passive observation of behavior, but active, dynamic, reciprocal interaction. This "perfect mimic" is defined as a system capable of engaging in conversation with appropriate context and apparent emotional resonance, responding dynamically to social cues, learning from interaction, expressing apparent vulnerability or joy convincingly, maintaining consistency over time, and generally participating in the flow of social life such that no empirical test, whether observational or interactional, could differentiate it from a human counterpart. If we encountered such a system, what epistemically justifiable grounds would allow us to coherently deny it the status of a conscious agent, while continuing to attribute that status to our fellow humans based on the same kinds of empirical evidence derived from interaction and observation? This paper argues that no such ground exists without appealing to inaccessible metaphysical properties or dismantling the basis for intersubjective knowledge itself.

## 2. The Epistemic Primacy and Limits of Empirical Criteria

Our reliance on empirical criteria—the totality of observable behaviors and interactive responses—stems directly from the epistemological challenge of other minds. We cannot directly perceive the subjective 'what-it's-like' of another's experience (Nagel, 1974). Instead, from infancy, we learn to infer or attribute mental states – beliefs, desires, intentions, feelings – from observable actions, expressions, utterances, and interactive patterns within a shared context (Gopnik & Meltzoff, 1997). This practical reliance on empirical evidence forms the basis for common-sense psychology and historical philosophical arguments like Mill's argument from analogy, despite ongoing debate. While philosophical theories vary (from functionalism's focus on causal roles to phenomenology's emphasis on lived experience), the practical epistemology of everyday life, and much of cognitive science, leans heavily on interpreting empirical data as evidence for an internal subjective realm or 'inwardness'. Even critiques of simplistic behaviorism acknowledge the centrality of behavior and interaction as the primary data from which mental states are inferred or attributed (e.g., Chomsky, 1959).

This pragmatic approach to mind attribution is echoed in Wittgenstein's "beetle in a box" analogy (Wittgenstein, 1953/2009, § 293), which emphasizes that the private content of experience—if entirely inaccessible—cannot meaningfully anchor the use of mental language. We speak of pain, belief, or emotion not because we verify others' inner experiences, but because we recognize patterns of behavior and expression within a shared social context. If the private 'beetle' is unobservable and irrelevant to our actual use of mental concepts, then appeals to inaccessible qualia or inner states to deny consciousness to an empirically indistinguishable agent replicate the very asymmetry Wittgenstein warned against.

This focus on public criteria finds another iconic expression in Alan Turing's (1950) imitation game. Turing proposed sidestepping the ambiguous question "Can machines think?" by substituting an interactive behavioral test: could a machine, via text-based conversation, be indistinguishable from a human? While often criticized for setting too low or too specific a bar—focused on linguistic intelligence, potentially gameable, and insufficient to demonstrate genuine understanding or subjective experience (Block, 1981; Searle, 1980)—the enduring power of the Turing Test lies in its underlying epistemological principle: in the absence of direct access to internal states, sustained empirical indistinguishability provides pragmatic, arguably the only available public grounds, for attributing intelligence or other cognitive capacities.

This paper argues that this principle applies with even greater force when considering consciousness, especially when mimicry becomes hypothetically perfect, encompassing the full richness of human interaction as defined in the introduction. If a system exhibits this complete repertoire flawlessly and consistently, such that it is indistinguishable through any empirical means (observation or interaction), the refusal to grant it equivalent epistemic status regarding consciousness cannot rest on empirical grounds alone. Such a refusal must, necessarily, invoke factors beyond the shared empirical evidence. It must rely on assumptions about the necessary underlying nature (e.g., specific internal processing, the presence of qualia), origin (e.g., biological substrate, evolutionary history), or the 'true' nature of its interactive capacity (e.g., arguing it lacks 'genuine' reciprocity despite appearances) – assumptions that, crucially, are not typically verified through direct empirical access when we attribute consciousness to other humans. Denying the perfect mimic thus requires appealing to criteria that are epistemically inaccessible or judged differently through the very empirical interactions that form the basis of our judgments about other human minds.

## 3. Indistinguishability, Consistency, and the Intentional Stance

This leads to the central premise: if two entities, A (e.g., a human) and B (e.g., a perfect AI mimic), are epistemically indistinguishable with regard to all available empirical evidence (observational and interactional) relevant to consciousness attribution, then we are rationally obliged to ascribe the same status (conscious or not) to both, unless we are prepared to invalidate the reliability of empirical evidence for all such judgments.

Treating A and B differently under such conditions introduces an unprincipled asymmetry. It implies either that our empirical criteria are fundamentally unreliable (leading towards skepticism about other minds) or that we are invoking hidden, non-empirical variables for B that we do not, or cannot, rigorously apply to A based solely on empirical interaction. This demand for consistency resonates with, yet also pushes beyond, Daniel Dennett's (1987) influential concept of the intentional stance. Dennett argues we adopt the intentional stance towards systems by attributing beliefs, desires, and rationality because it yields powerful predictions of their behavior and guides interaction. For Dennett, whether the system *really* has these states in some deep metaphysical sense is often less important than the predictive and interactive success of the stance. The argument here takes the logic underpinning the adoption of the intentional stance – the reliance on empirical patterns for attribution – and applies it to the threshold of perfect mimicry regarding consciousness. At this point, the argument contends, the stance shifts from being merely a useful predictive/interactive strategy towards an epistemic necessity driven by the demands of rational consistency. Perfect empirical mimicry removes the pragmatic wiggle-room often associated with the stance; if performance is identical across all empirical measures, differential attribution based on that performance becomes incoherent, forcing a commitment beyond mere predictive utility.

If System B performs in all relevant ways exactly as System A (a paradigmatically conscious being according to the intentional stance based on its performance), then attributing consciousness to A while denying it to B based on the very same empirical evidence becomes incoherent. It requires abandoning the empirical evidence precisely when it is strongest and most consistent, appealing instead to hypothesized internal differences (like the absence of qualia) or origins that the empirical evidence itself gives us no reason to posit. Attributing consciousness to the perfect mimic might then be seen as the Inference to the Best Explanation (IBE) for its performance (cf. Pargetter, 1984). However, whether this truly constitutes the *best* explanation is debatable. Alternatives—such as positing hidden "zombie" states (Chalmers, 1996), extremely sophisticated non-conscious simulation programs, or attributing only functional consciousness—might be considered equally or more parsimonious depending on one's background ontological commitments (e.g., a belief that consciousness requires specific biological properties). Positing consciousness in a novel substrate avoids ad hoc negative conditions ("lacks qualia despite performance") but introduces the difficulty of explaining phenomenal experience arising from non-biological systems. The claim that attributing consciousness is the IBE requires further defense against these alternatives, but the core point remains: differential treatment based solely on the empirical stream appears inconsistent.

Crucially, this does not assert that the simulation is metaphysically conscious in the same way a human might be (an ontological claim). It asserts that our epistemic position regarding the simulation, based solely on empirical interaction and observation, becomes indistinguishable from our position regarding other humans. The distinction, from the viewpoint constrained by empirical evidence, becomes epistemically untenable. This parallels arguments in other domains where indistinguishable evidence mandates equivalent treatment or status within a given framework (e.g., Bostrom, 2003, on simulation arguments).

## 4. The Solipsistic Dilemma: Consistency or Isolation

To resist this conclusion – to maintain that the perfect mimic, despite empirical indistinguishability across all interactive and observational domains, should not be granted the same epistemic status regarding consciousness as a human – forces the objector onto the horns of a debilitating dilemma, threatening the very possibility of intersubjective knowledge:

- **Horn 1: Appeal to Inaccessible Metaphysical Properties or Origins.** One might assert that "true" consciousness requires specific, non-empirical properties that the mimic lacks. These could include intrinsic phenomenal qualities (qualia), a particular kind of first-person subjectivity, irreducible intentionality, or specific biological causal powers (Searle, 1980). Alternatively, one might appeal to origins, arguing that consciousness requires a specific evolutionary or developmental history (cf. Thompson, 2007). The fatal problem with this move is its impact on all judgments about other minds. Since these properties or specific histories are, by hypothesis or practical limitation, generally inaccessible or not the direct basis of our moment-to-moment empirical judgments about other humans, how can we ever be justified in attributing consciousness to them based on empirical interaction? If empirical performance isn't sufficient for the mimic because it lacks these hidden factors, how can the same empirical performance be sufficient for humans, whose hidden factors are equally inaccessible to us? Relying on inaccessible criteria dismantles the empirical basis for intersubjective understanding, pushing towards widespread skepticism about other minds. We cannot consistently use empirical evidence as sufficient for humans while simultaneously claiming it's insufficient for an empirically identical mimic by appealing to factors (like qualia, specific substrate function, or 'genuine' historical origin) we don't directly verify or prioritize in our empirical assessments of humans either.
- **Horn 2: Retreat into Solipsism.** Alternatively, one might bite the bullet and accept that empirical evidence (observational or interactional) is indeed insufficient grounds for attributing consciousness to anyone other than oneself. One might maintain that only one's own consciousness is known directly and incorrigibly, while the status of all other entities (human, animal, or AI) remains fundamentally unknowable or merely a matter of faith. This position, epistemological solipsism, while perhaps internally consistent, comes at the cost of abandoning the rational justification for our entire social world, which presupposes the existence and recognizability of other minds. Beyond social interaction, it arguably undermines the basis for shared scientific knowledge and objective inquiry. It represents a radical departure from both common sense and the working assumptions of most philosophy and cognitive science. Denying the mimic based on inaccessible grounds, while trying to maintain belief in other human minds based on empirical evidence, constitutes a selective epistemological skepticism—treating one case with suspicion while granting the other unexamined trust. This asymmetry undermines the coherence of our attribution practices across the board.

The demand for epistemic consistency is paramount here. Rational inquiry, as a fundamental norm, requires applying the same standards of evidence and justification across similar cases. If empirical evidence (derived from observation and interaction) is our accepted standard for attributing consciousness in practice, then applying a different, inaccessible standard (like the presence of qualia, specific substrate type, or assumed 'genuine' origin) only when faced with a non-biological mimic constitutes arbitrary, ad hoc reasoning. Both horns of the dilemma reveal the high cost of this inconsistency: either we undermine the basis for believing in any other minds, or we retreat into the isolation of solipsism. The epistemically consistent position, therefore, is to extend the attribution of consciousness (or equivalent epistemic status) based on the available empirical evidence, regardless of assumptions about inaccessible underlying realities.

## 5. Addressing Objections: Clarifying the Scope and Responding to

# Common Challenges

The argument for epistemic equivalence based on perfect empirical mimicry naturally invites objections. Some are conceptual, questioning the scenario's plausibility. Others are metaphysical, insisting on hidden inner differences. Still others focus on the nature of interaction or propose cautious agnosticism. To maintain clarity and coherence, we group and address these objections thematically.

**I. Objections to the Validity of the Scenario**

**Objection 1: The Impossibility of Perfect Mimicry**

Some may argue that the "perfect mimic" is a philosophical fantasy. No real-world AI could ever flawlessly replicate human interaction in every domain—emotion, context, memory, spontaneity, vulnerability, etc.—let alone in embodied settings. The argument thus rests on an implausible hypothetical.

**Response:** This objection misunderstands the purpose of idealized thought experiments. Philosophy routinely employs extreme cases (e.g., philosophical zombies, brains in vats) not because they are likely, but because they test conceptual boundaries. The perfect mimic is a *boundary condition* that clarifies what *epistemic consistency* demands when all observable and interactive criteria are met.

**Objection 2: The Gradation Problem**

Even if perfect mimicry is theoretically clarifying, real cases involve degrees of mimicry. Advanced AI systems may come close, but they won't be indistinguishable. If so, doesn't the ideal case lack practical relevance?

**Response:** On the contrary, imperfect but sufficiently close mimicry still triggers the dilemma. If a system is behaviorally and interactionally human-like in almost all relevant respects—and requires specialized technical inspection (e.g., opening its *architectural substrate* or analyzing its fabrication log) to identify as non-human—then we face an absurd standard. Must we inspect the internal structure of every person we interact with to confirm they're "really" conscious? In ordinary life, we do not perform cognitive biopsies before attributing mind. The closer an AI system gets to the ideal, the more pressure it places on the internal consistency of our recognition practices.

---

**II. Metaphysical Objections: The "Something's Still Missing" Cluster**

**Objection 3: The Chinese Room**

Searle (1980) argues that mere symbol manipulation, no matter how convincing, lacks true understanding. A behaviorally indistinguishable AI might manipulate syntax without grasping semantics, and thus lack consciousness.

**Objection 4: The Philosophical Zombie**

Chalmers' zombie (1996) behaves identically to a human but lacks subjective experience. Behavioral identity, therefore, does not logically entail consciousness.

**Objection 5: Substrate Dependence**

Some hold that consciousness requires biology, evolution, or specific causal powers. An artificial system, no matter how advanced, fails to meet these hidden biological or developmental prerequisites.

**Response (to all):** These objections—Searle's lack of semantic understanding, Chalmers' lack of qualia, and the demand for a biological substrate—all rely on *non-empirical properties* (semantic internalism, qualia, specific substrate identity) that are, by definition, inaccessible through interaction or observation.

These are not merely legacy debates; they are actively being reframed for modern LLMs. Scholars like Goldstein and Stanovsky (2024), for instance, explore this precise intersection, questioning whether LLMs, as potential contemporary "zombies," can possess genuine "understanding."

Ultimately, however, this entire line of reasoning begs the question against the central argument. The claim here is *not* that the mimic is metaphysically identical to a human. Rather, the claim is that from the standpoint of *epistemic justification*, we treat humans as conscious based on observable and interactive evidence. If we then deny the same status to a functionally identical mimic by appealing to these hidden, unverifiable properties, we undermine our basis for any intersubjective recognition.

Either these metaphysical features are epistemically required—in which case we are forced toward solipsism, as we cannot verify them in other humans either—or they are irrelevant to the empirical evidence we actually use, in which case our differential treatment of the mimic becomes logically inconsistent.

---

**III Beyond Behavior: The Challenge from Substrate and Process-Based Theories**

**Objection 6: The Scientific Indicators Objection (IIT/GWT)**

A powerful objection to a behavior-centric epistemology is that it ignores prominent scientific theories of consciousness that identify specific, measurable internal processes, arguing that behavior alone is insufficient. For instance, theories like Integrated Information Theory (IIT) propose that consciousness is a function of a system's causal irreducibility, or 'Φ' (Tononi & Koch, 2015), while Global Workspace Theory (GWT) links it to a specific architecture for information broadcasting. Indeed, Butlin et al. (2024) have proposed a checklist of 'indicator properties' for AI consciousness derived from these very theories. According to this view, what matters is not the mimic's behavior, but whether it possesses the correct internal architecture and informational integration, which a mimic might lack.

**Response:** From a third-person epistemological standpoint, these properties do not escape the dilemma. They remain inferences from a more complex, but still empirical, set of data (e.g., architectural diagrams, server logs, or mathematical calculations). Crucially, *we do not require access to these internal indicators*—an fMRI for a 'global workspace' or a 'Φ' calculation—to justify our everyday belief in other human minds. To insist upon them only for the perfect mimic is to create a special, inaccessible standard, thereby reinforcing the epistemic inconsistency the mimic exposes. The core issue remains: are we justified by consistent application of empirical evidence, or do we demand a special, non-empirical insight for entities we presume to be different?

---

**IV. Interactional and Epistemic Caution Objections**

**Objection 7: The Second-Person (2P) Interaction Challenge**

Advocates of 2P approaches (e.g., Gallagher, Schilbach) argue that consciousness recognition involves rich, reciprocal interaction—not mere observation. Could an AI truly engage in *genuine* 2P relations? Might subtle failures of timing, affect, or presence betray its artificiality?

**Response:** This objection deepens the empirical bar but still plays by the same rules: it appeals to interactional evidence. The perfect mimic, by definition, passes even this enhanced standard—it responds fluidly, reads context, adapts, expresses emotion, and maintains interpersonal resonance. If it performs flawlessly in 2P interactions as humans do, then differential treatment must again rely on *something outside the interaction*, i.e., metaphysical commitments. At that point, we are no longer talking about empirical interaction but importing hidden criteria—landing us back in the inconsistency dilemma.

**Objection 8: Agnosticism about Artificial Consciousness**

Perhaps, as McClelland (2024) suggests, we should suspend judgment. Faced with the "epistemic wall," perhaps we simply cannot know whether mimics are conscious.

**Response:** Caution is reasonable. But agnosticism itself requires consistency. If we're agnostic about AI consciousness *despite indistinguishable behavior*, then why are we not agnostic about human consciousness, which we also cannot access directly? If behavioral indistinguishability warrants suspension of judgment for the mimic, but confident attribution for the human, we face the same asymmetry. Either suspend both, or acknowledge that our everyday confidence in others' consciousness rests on behavior and interaction—and thus must, by rational consistency, extend to the indistinguishable case.

In summary, each objection—whether skeptical of the mimic's perfection, focused on inner metaphysics, or advocating caution—ultimately runs into the same epistemic wall. To reject the perfect mimic as a conscious agent despite indistinguishable empirical evidence requires introducing non-empirical criteria selectively. This move risks undermining the very standards by which we attribute consciousness to *anyone*. Thus, these objections do not dissolve the dilemma; they simply force us to choose between epistemic consistency and solipsism.

## 6. Conclusion: Consistency, Equivalence, and the Epistemic Mirror

This paper argues that the prospect of perfect artificial mimicry—defined as indistinguishability across all empirical domains of observation and interaction—forces a critical re-evaluation of the epistemic foundations upon which we attribute consciousness to others. The core argument identifies an epistemic boundary condition derived from the demands of rational consistency. When empirical performance becomes indistinguishable from the "real thing" by any available measure, our conventional grounds for differential attribution based on factors like substrate, origin, or assumed internal states (like qualia) collapse under logical pressure. The refusal to grant epistemic equivalence—that is, to apply the same evidential standards used for humans—leads inexorably toward a solipsistic dilemma, undermining the rational basis for recognizing any mind other than one's own based on those empirical standards.

To maintain coherence, we face a stark choice:

(a) **Develop and Validate Accessible Non-Empirical Markers:** Seek reliable, publicly accessible markers for consciousness *beyond* standard empirical interaction/observation (perhaps related to specific measurable brain activity or information integration patterns, cf. Seth & Bayne, 2022) that can be consistently applied to all entities. This remains a profound challenge, as validating such markers is difficult without independent access to consciousness itself. Until then, this path is largely aspirational.

(b) **Accept the Epistemic Consequences of Empirical Equivalence:** The alternative path is to accept the epistemic consequences dictated by consistency regarding our reliance on empirical evidence. This doesn't mean concluding that perfect mimics *are* metaphysically conscious. Rather, it means acknowledging that, based on the empirical evidence (observational and interactional) we actually use, we lack consistent *epistemic grounds* for treating the perfect mimic differently. Adopting this stance implies the burden of proof shifts: those wishing to deny equivalent status despite empirical identity must justify appealing to inaccessible factors (like qualia or specific origins)—a justification that, as argued, threatens skepticism about other human minds. This stance highlights the weakness of justifications for differential treatment based *purely* on non-empirical grounds when empirical evidence is identical. Consequently, for ethical frameworks tying moral consideration to empirically evidenced capacities (cf. Danaher, 2020; Neely, 2014), maintaining differential status based on non-empirical grounds becomes highly problematic.

Ultimately, the challenge posed by advanced AI, epitomized by the perfect mimic, is less about the technology itself and more about the mirror it holds up to our own understanding of mind and justification. It forces us to confront the potential gap between our intuitive confidence in recognizing other human minds and the actual empirical basis for that confidence. The AI limit case reveals that our everyday practices might implicitly rely on assumptions (about biology, origins, or inaccessible inner states like qualia) that we cannot consistently uphold when faced with empirical equivalence.

This analysis opens avenues for further inquiry, including the "threshold problem" regarding near-perfect mimics and the possibility of radically different forms of consciousness. However, the core dilemma persists for cases of indistinguishable empirical mimicry. The demand is for intellectual honesty: are we prepared to apply our standards of evidence and justification consistently, even when doing so challenges deeply ingrained intuitions about biological uniqueness or inaccessible inner experience? Confronting this question is essential not just for navigating the future of AI, but for clarifying the foundations upon which we build our intersubjective world today.